\documentclass{article}

\usepackage{microtype}
\usepackage{graphicx}
\usepackage{subcaption}
\usepackage{booktabs}
\usepackage{multirow}
\usepackage{amsmath}
\usepackage{amssymb}
\usepackage{mathtools}
\usepackage{algorithm}
\usepackage{tikz}
\usepackage{pgfplots}
\pgfplotsset{compat=1.18}
\usepgfplotslibrary{groupplots}
\usepackage{algorithmic}
\usepackage{hyperref}
\usepackage[capitalize,noabbrev]{cleveref}
\usepackage{multirow}

\usepackage[preprint]{icml2026}

\newcommand{\zres}{z_{\mathrm{res}}}

\newcommand{\tgt}{c_{\mathrm{tgt}}}
\newcommand{\flow}{v_\theta}

\newcommand{\norm}[1]{\left\lVert#1\right\rVert}

\icmltitlerunning{Residual-Space Evolutionary Optimization via Flow-based Generative Models}

\begin{document}

\twocolumn[
\icmltitle{Residual-Space Evolutionary Optimization via Flow-based Generative Models}

 \icmlsetsymbol{equal}{*}

  \begin{icmlauthorlist}
    \icmlauthor{Zhuo Cao}{fzj}
    \icmlauthor{Lena Krieger}{fzj,comp}
    \icmlauthor{Fernanda Nader}{fzj}
    \icmlauthor{Xuan Zhao}{fzj}
    \icmlauthor{Hanno Scharr}{fzj}
    \icmlauthor{Ira Assent}{fzj,sch}
  \end{icmlauthorlist}

  \icmlaffiliation{fzj}{IAS-8, Forschungszentrum Jülich, Germany}
  \icmlaffiliation{comp}{LMU Munich, Munich Center for Machine Learning (MCML), Germany}
  \icmlaffiliation{sch}{Department of Computer Science, Aarhus University, Denmark}

  \icmlcorrespondingauthor{Zhuo Cao}{z.cao@fz-juelich.de}

\icmlkeywords{data optimization, conditional flow matching, counterfactual explanations, evolutionary algorithms}
\vskip 0.3in
]

\printAffiliationsAndNotice{}

\begin{abstract}
Data editing with generative methods typically requires differentiable objectives and gradient-based search. However, these assumptions break down in flow-based settings, where edits are performed through forward and backward integration and often involve non-differentiable or black-box objectives.
We introduce \textbf{residual-space evolutionary optimization}, a model-agnostic framework that addresses this gap by combining flow-based generative editing with evolutionary algorithms. 
Building on the observation that conditional flow matching (CFM) can disentangle condition-controlled factors from instance-specific residuals, our framework directly operates in residual space and 
separates two complementary search regimes: \emph{self-pollination} performs local exploitation through feature-preserving residual refinement, and \emph{cross-pollination} promotes broader exploration by recombining residuals across heterogeneous samples. 
As a proof of concept, we validate on MorphoMNIST, a benchmark dataset for counterfactual generation, and on crystal data, demonstrating
that this exploration--exploitation decomposition provides a useful mechanism for balancing target alignment, instance preservation, and diversity, and extends beyond images to real-world scientific domains.
\end{abstract}

\section{Introduction}

Controllable data editing, i.e., modifying targeted attributes while preserving instance-specific structure, is a core operation in machine learning, in applications ranging from counterfactual explanations to data augmentation. Beyond images, it is equally important in scientific domains such as drug discovery, crystal structure prediction, and materials optimization, where controllable edits can steer valid samples toward desired functional properties.
Most existing approaches treat editing as gradient-based optimization, implicitly assuming that objectives are differentiable and that the generative pipeline is fully transparent.
A broad line of work in the image domain, including feature visualization and network dissection
\citep{mahendran2015understanding,olah2017feature,carter2019activation,bau2017network,bau2019gan,selvaraju2020grad},
style transfer and image-to-image translation
\citep{gatys2016image,zhu2017unpaired,isola2017image,park2020contrastive},
GAN inversion and latent editing
\citep{abdal2019image2stylegan,shen2020interpreting,harkonen2020ganspace,patashnik2021styleclip,roich2022pivotal,pan2023drag},
and diffusion-based image editing and controllable generation
\citep{dhariwal2021diffusion,meng2022sdedit,hertz2023prompt,mokady2023null,brooks2023instructpix2pix,zhang2023adding,parmar2023zero,mou2024t2i}, share this assumption, that treating editing as optimization problems over pixels, features, or latent variables is feasible.
This assumption does not hold in flow-based generative editing, where edits are implemented through forward and backward numerical integration 
and objectives are often non-differentiable or black-box. 
Recent work shows that conditional flow matching (CFM) disentangles condition-controlled factors from instance-specific residual information \citep{li2024return,cao2025leapfactual}, enabling iterative editing through repeated integration. This iterative mechanism is naturally well-suited to evolutionary algorithms \citep{holland1975adaptation,goldberg1989genetic,back1996evolutionary,eiben2015introduction,hansen2001completely}, which operate through repeated proposal, evaluation, and refinement, enabling residual-space edits to act as genotype-like variations that can be selected to optimize target properties.

\begin{figure}[ht]
  \centering
  \includegraphics[width=0.99\columnwidth]{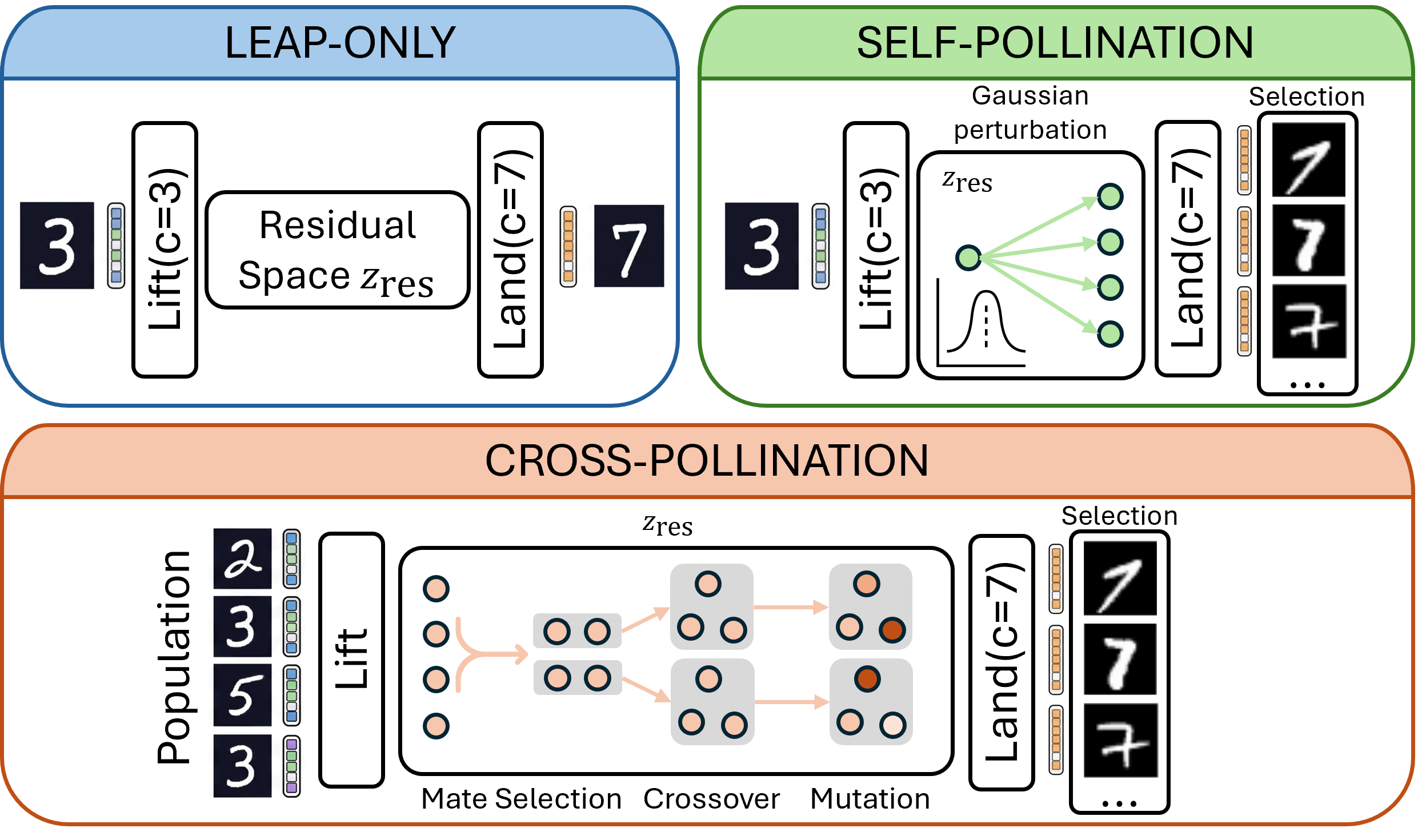}
  \caption{\textbf{Comparison of leap-only, self-pollination, and cross-pollination.} Colored boxes visualize the individual methods.}
  \label{fig:qualitative}
\end{figure}

We propose \emph{residual-space evolutionary optimization}, a model-agnostic framework that combines flow-based generative editing with evolutionary algorithms. Given a fixed conditional generator, our method maps data into residual states, edits these states through mutation and crossover, and decodes the resulting candidates under a target condition. Selection is then performed using task-specific criteria such as target validity, instance preservation, feature control, or diversity (see \Cref{fig:qualitative}), without requiring gradient information from the generator. Thus, the method acts as a lightweight optimization layer on top of an existing generator, rather than a new generative model training objective. 

A central perspective of our framework is that residual-space evolution factorizes the classical exploration--exploitation trade-off into two pollination mechanisms. Self-pollination exploits the local residual neighborhood of an existing sample, making it suitable for refinement problems where preserving the source instance is important. Cross-pollination explores a broader residual search space by recombining information across heterogeneous samples, which can help discover diverse candidates and mitigate premature convergence to a local optimum. Importantly, we do not claim that cross-pollination guarantees a global optimum; rather, it provides a mechanism for increasing coverage of the target-conditioned solution space before selection.

We instantiate the framework based on the existing work LeapFactual \citep{cao2025leapfactual} and evaluate it on MorphoMNIST \citep{castro2019morphomnist} as a controlled image-editing testbed. Although images provide a convenient visualization domain, the framework is not image-specific and can apply to any conditional data editing setting with an editable latent or residual representation. We further validate the framework on the Wyckoff inorganic crystal generator (WyCryst) \citep{zhu2024wycryst}, demonstrating applicability beyond the image domain to real-world scientific data. 
Our results demonstrate that residual states exposed by flow-based generative editors constitute effective search spaces for controlled editing, with the exploration-exploitation decomposition providing explicit mechanisms for balancing target alignment, instance preservation, and diversity.

\section{Method}
\subsection{Preliminaries}
\paragraph{Evolutionary Algorithms.} Evolutionary algorithms are population-based optimization methods inspired by natural selection. Given a population of candidate solutions, they iteratively generate new candidates through stochastic variation operators, such as \emph{mutation} and \emph{crossover}, and retain promising candidates through \emph{selection}. Mutation perturbs an individual candidate to explore its local neighborhood, while crossover recombines information from multiple candidates to produce new offspring. Selection then evaluates candidates according to a task-specific fitness function and keeps those that best satisfy the desired objective. In our framework,  residual states serve as candidate representation, flow-based edits act as variation operators, and task-specific criteria define the fitness function.

\paragraph{LeapFactual.} Our framework builds on the flow-based editing formulation of \citet{cao2025leapfactual}, which we briefly review. Let $x$ denote an input image, $z=E(x)$ its autoencoder latent representation, $\hat{c}=f(x)$ its predicted source class, and $\tgt$ a user-specified target class. We assume a single shared conditional flow model $\flow(z_t,t,c)$ trained with class conditions. At editing time, the same flow is used in two integration directions.

The source-conditioned reverse integration, referred to as \emph{lifting}, removes class-related information from the latent $z$ 
\begin{equation}
  \zres = \operatorname{Lift}(z, \hat{c}) ,
\end{equation}
returning the residual state $\zres$,
where the flow is integrated backward from $t=1$ to $t=0$. The target-conditioned forward integration, referred to as \emph{landing}, reconstructs a complete latent $z'$ under the desired target condition $\tgt$:
\begin{equation}
  z' = \operatorname{Land}(\zres, \tgt) ,
\end{equation}
where the flow is integrated forward from $t=0$ to $t=1$. The combination of a \emph{Lift} and a \emph{Land} operation forms a \emph{Leap}, and the edited image is then obtained as $x'=D(z')$ using the autoencoder's decoder $D$.

\paragraph{Design Principle.}
In the experiments below, we use this formulation as a concrete instantiation of a broader residual-space optimization principle. All search operations are performed in $\zres$, rather than in image space or in the latent space of the autoencoder, preserving a clean separation: class-related information is controlled by the source and target conditions, while instance-specific residual variation is manipulated by the search procedure. Although source and target conditions may be identical, we show empirically that allowing class changes enables more data-efficient use of residual information across instances.

\subsection{Residual-Space Evolutionary Optimization}
\label{sec:method}

We introduce an evolutionary layer on top of a frozen conditional flow model, treating the residual state $\zres$ as the searchable genome of a sample, while leaving the conditional flow model responsible for imposing the source and target conditions through \emph{Lift} and \emph{Land}. This design separates two roles: the flow model controls semantic conditioning, whereas the evolutionary layer searches over instance-specific variation.

Depending on how the residual population is constructed, this framework leads to two complementary search regimes. \emph{Self-pollination} instantiates local exploitation: it starts from a single residual and uses mutation to refine candidates within the neighborhood of an existing solution. 
\emph{Cross-pollination} instantiates broader exploration: it starts from multiple residuals and recombines them through crossover, allowing residual information from heterogeneous sources to generate candidates in different regions of the target-conditioned space. Pseudocode is in Appendix \ref{app:algorithm}.

\paragraph{Self-pollination.}
For a single input $x$, self-pollination first computes its residual state through the source-conditioned \emph{Lift} operation: $\zres = \operatorname{Lift}(E(x), \hat{c})$.
It then constructs a child pool with size $m$ by sampling perturbed residuals,
\begin{equation}
  \tilde{z}^{(m)}_{\mathrm{res}} = \zres + \epsilon^{(m)}, \qquad
  \epsilon^{(m)} \sim \mathcal{N}(0, \sigma^2 I),
\end{equation}
where alternative perturbations, such as feature swapping, can also be used when residual dimensions are treated as exchangeable genes. Each child residual is then \emph{landed} under the target condition and decoded: $x'^{(m)} = D(\operatorname{Land}(\tilde{z}^{(m)}_{\mathrm{res}}, \tgt))$.

Selection keeps the best candidates according to a user-defined fitness score. 
Thus, self-pollination performs local residual exploitation around one input and is mainly used for feature-preserving refinement. This makes it a natural fit for attractive objectives such as counterfactual explanation \citep{dombrowski2023diffeomorphic,10.1007/978-3-030-01249-6_41,singla2019explanation,nemirovsky2020countergan,kim2021counterfactual,hvilshoj2021ecinn,cao2025galaxy,cao2025leapfactual}, where the search should converge toward a target condition without unnecessarily drifting away from the source instance.

\paragraph{Cross-pollination.}
For a population $\{x_i\}_{i=1}^{N}$, cross-pollination first computes source-conditioned residuals
\begin{equation}
  z_{\mathrm{res},i}
  =
  \operatorname{Lift}(E(x_i), \hat{c}_i),
  \qquad i=1,\dots,N .
\end{equation}
Each residual will then be paired by a partner residual and recombined through crossover 
\begin{equation}
    \tilde{z}_{\mathrm{res,i}}^{(m)}=\mathrm{Crossover}(z_{\mathrm{res},i}, z_{\mathrm{res},j}^{(m)}, \alpha) + \epsilon^{(m)},
\end{equation}
where $\alpha$ controls the contribution of each parent, and $\epsilon^{(m)}$ is an optional mutation term. Different crossover mechanisms can be applied. The details can be found in \Cref{app:algorithm}. 

The resulting child residuals are landed under the same target condition and decoded. Selection again keeps the top-k candidates according to the defined fitness score. Unlike self-pollination, which preserves the identity of one source sample, cross-pollination uses residual diversity across multiple sources to explore a broader target-conditioned search space. Therefore, diversity is important to prevent early convergence. Optionally, advanced selection mechanisms, such as tournament and diverse greedy selection \citep{5949793,10900033,10730362}, can be applied. In our experiments, we demonstrate that the simple top-k mechanism works well since the diversity induced by cross-pollination prevents premature convergence (\Cref{sec:experiments}).

\section{Domain Modelling}
\label{sec:experiments}
In the following we introduce the experimental setups for both image  (Sec. \ref{subsec:img}) and scientific domain (Sec. \ref{subsec:crystal}).
\subsection{Image Domain: MorphoMNIST}\label{subsec:img}
We use MorphoMNIST \citep{castro2019morphomnist} as a studied domain because it provides interpretable scalar attributes, including morphological attributes such as thickness, slant, and width. 
Our model stack consists of an image classifier, a VAE-style latent encoder--decoder, and a single class-conditional CFM model used for source-conditioned lifting and target-conditioned landing. The framework supports arbitrary source-to-target digit pairs.

Our study is designed to demonstrate the complementary roles of the two proposed variants through an exploration--exploitation decomposition. 

\paragraph{Self-pollination.}
For self-pollination, we test whether residual-space mutation can refine a leap-only edit while better preserving instance-specific information from the input sample. Specifically, we select edits that maximize source similarity while encouraging target-class confidence:
\begin{equation}
    S_{\mathrm{self}}(x',x)
    =
    \mathrm{sim}(x',x)
    + \lambda \, p_{\mathrm{clf}}(\tgt \mid x'),
\end{equation}
where \(\mathrm{sim}(x',x)\) is an image similarity measure defined in \cref{app:definitions}, \(p_{\mathrm{clf}}(\tgt \mid x')=\mathrm{softmax}(C(x'))_{\tgt}\), and \(\lambda>0\) controls the trade-off between target confidence and source preservation. 

\paragraph{Cross-pollination.}
For cross-pollination, we shift the objective from instance preservation to feature exploration. Here, we test whether residual information from non-target classes can serve as genetic material for target-conditioned synthesis. Rather than optimizing only over samples that already belong to the target digit, cross-pollination supports broader exploration by collecting residual ``genes'' from a mixed-source population and landing all offspring under the same target condition. Specifically, we select edits that maximize digit thickness, measured using the MorphoMNIST \citep{castro2019morphomnist}, illustrating that the framework supports non-differentiable, black box fitness functions, while encouraging target-class confidence:
\begin{equation}
    S_{\mathrm{cross}}(x')
    =
    \mathrm{thickness}(x')
    + \lambda \, p_{\mathrm{clf}}(\tgt \mid x'),
\end{equation}
where $\mathrm{thickness}(x')$ is the morphological thickness and $p_{\mathrm{clf}}(\tgt \mid x')=\mathrm{softmax}(C(x'))_{\tgt}$.
More details can be found in Appendix \ref{app:definitions} and \ref{app:experiment}.

\paragraph{Baselines.}
The two variants differ fundamentally in their starting population: self-pollination always begins from a single source sample, whereas cross-pollination begins from a set of samples whose composition defines the search regime.
For self-pollination, the main baseline is a leap-only method without residual-space evolution. 
For cross-pollination, the baseline uses a homogeneous population drawn entirely from the target class, while the proposed diverse cross-pollination draws from a mixed-source population and decodes every child under the same target condition, allowing residual information from other classes to contribute to target-conditioned generation.

\paragraph{Metrics.}
For self-pollination, we report validity and similarity. Validity is the fraction of generated images classified as the target class, reported for both variants. Similarity is measured as mean RMSE between the generated and the input image.
For cross-pollination, we report validity, feature value, and diversity. 
Feature value reports mean digit thickness in the top 95th percentile of the population, measuring the framework's ability to produce high-attribute-value candidates under the target condition. Diversity is measured as the angular distance between normalized flattened images from the input and generated populations.
More details can be found in Appendix \ref{app:definitions}.


\subsection{Scientific Domain: Crystal Data}\label{subsec:crystal}
We conduct the experiment using WyCryst \citep{zhu2024wycryst}, a VAE-based model for encoding and decoding material structures. The dataset to generate the latent space was provided by the authors via the project repository, originally queried from the Materials Project database (v.2023.7.4) \cite{jain2013materials} and containing 66,643 ternary inorganic compounds. After filtering to structures with at most 20 atoms per unit cell, formation energy $\leq 1$~eV/atom, and energy above the convex hull $E_\mathrm{hull} < 0.1$~eV/atom, the working set comprises 28,318 crystal structures spanning 87 unique elements and all seven crystal systems: cubic, hexagonal, trigonal, tetragonal, orthorhombic, monoclinic, and triclinic. In the latent space learned by WyCryst, we train three additional models: a classifier that predicts the crystal system classes; a regressor that predicts the band gap, which serves as the scalar material property and optimization objective; and a conditional flow matching model. Unlike the image-domain setting, all components in this setting are trained and operate directly in the material latent space.

\paragraph{Cross-pollination.}

We apply cross-pollination to maximize the predicted band gap, steering structures toward wider-gap insulating phases, while simultaneously changing the crystal system of the material.

This allows the algorithm to reuse residual information, or “genetic” features, from materials belonging to different crystal systems, expanding the search space beyond within-class variations.
Specifically, we select edits that maximize band gap while encouraging crystal system confidence:
\begin{equation}
    S_{\mathrm{cross}}(x')
    =
    \mathrm{bandgap}(x')
    + \lambda \, p_{\mathrm{clf}}(\tgt \mid x'),
\end{equation}
where $ \mathrm{bandgap}(x')$ is predicted by the trained regressor and $p_{\mathrm{clf}}(\tgt \mid x')=\mathrm{softmax}(C(x'))_{\tgt}$.

\paragraph{Baselines.}
As with the image-domain experiments, cross-pollination begins from a set of samples whose composition defines the search regime. The baseline uses a homogeneous population drawn from the target crystal system. The diverse cross-pollination variant draws from a mixed-source population spanning all crystal systems, allowing residual information from structurally distinct materials to contribute to target-conditioned generation and to expand the search space beyond within-class variation.
\paragraph{Metrics.}
\textcolor{black}{We report validity, feature value, and diversity. Validity is the fraction of generated crystal classified as the target crystal. Feature value is the mean band gap among samples in the top 95th percentile of the population. Diversity is measured as the angular distance between normalized latent representations of the generated populations.}

\section{Results}\label{sec:results}

\paragraph{Self-pollination supports local exploitation.}
As shown in \cref{tab:overall_self_pollination}, self-pollination maintains near-perfect validity ($>0.99$) while consistently improving source-instance preservation. 
Specifically, self-pollination improves similarity by 3\% over the leap-only baseline, indicating that residual-space mutation refines the target-conditioned edit while better preserving input-specific structure.

\begin{figure}[ht]
  \centering
  \includegraphics[width=0.9\linewidth]{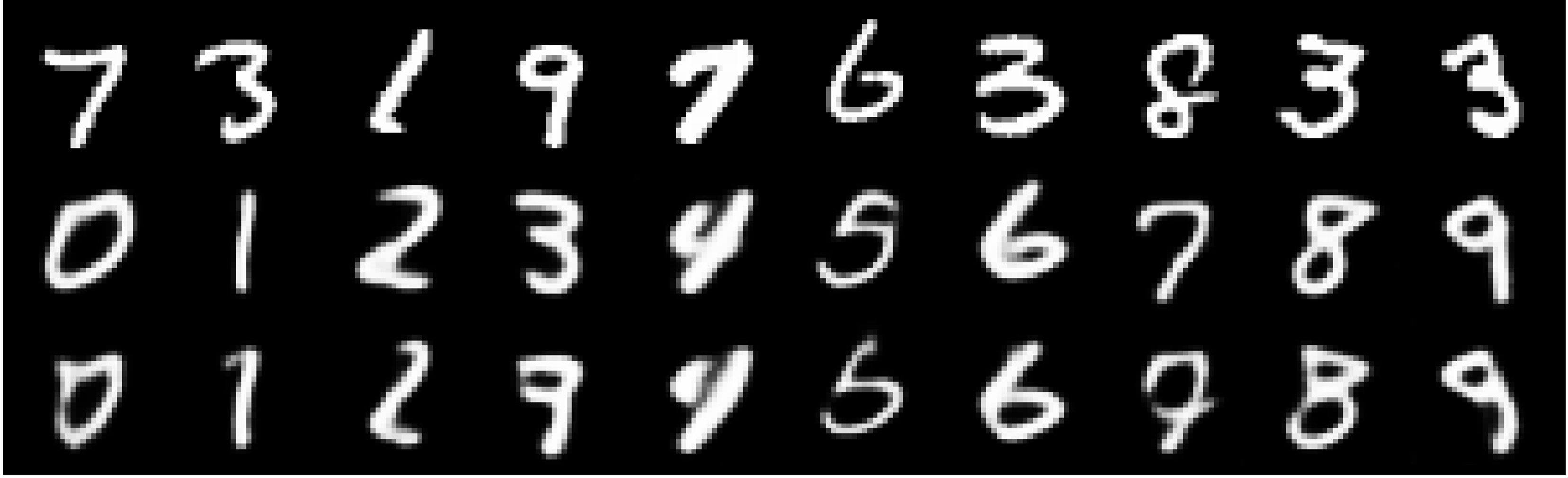}
  \caption{
  \textbf{Qualitative comparison of leap-only and self-pollination across target digits.} Self-pollination better preserves input-specific stroke style while achieving the target digit. Columns are target digits (0-9). Rows are input (top), leap-only (middle), and self-pollination (bottom).}
  \label{fig:self-demo}
\end{figure}

The qualitative examples in \cref{fig:self-demo} support this interpretation. While both leap-only and self-pollination successfully reach the desired target classes, self-pollination better preserves instance-level characteristics such as stroke thickness, slant, and writing style from the input images. A per-digit breakdown is provided in \cref{tab:per_digit_self_pollination} in the appendix, where self-pollination improves similarity for every target digit and maintains or improves validity across all digits. These results suggest that self-pollination acts as an effective local exploitation mechanism, improving source-instance preservation without sacrificing target validity.

\begin{table}[ht]
\centering
\caption{
Aggregated comparison between leap-only and self-pollination on MorphoMNIST.
Self-pollination is reported as the relative improvement over leap-only.
Results are shown as mean$_{\pm \mathrm{STE}}$ over three random seeds.
}
\label{tab:overall_self_pollination}
\resizebox{0.75\columnwidth}{!}{
\begin{tabular}{lcc}
\toprule
Method & Validity $\uparrow$ & Similarity $\uparrow$ \\
\midrule
Leap-only 
& $\phantom{+}0.9989_{\pm 0.0001}$ 
& $\phantom{+}0.9316_{\pm 0.0001}$ \\
Self-pollination 
& $+0.0011_{\pm 0.0001}$ 
& $+0.0310_{\pm 0.0001}$ \\
\bottomrule
\end{tabular}
}
\end{table}
\begin{table}[ht]
\centering
\caption{
Aggregated comparison of homogeneous vs. diverse cross-pollination.
Feature value and diversity differ by domain: MorphoMNIST reports optimized morphological feature and image-space diversity; crystal data reports predicted band gap and latent diversity.
Diverse is reported as the relative improvement over homogeneous.
The final generation results are shown as mean$_{\pm \mathrm{STE}}$ over all target classes/systems and three random seeds.
}
\label{tab:cross_pollination_aggregate}
\resizebox{\columnwidth}{!}{
\begin{tabular}{llccc}
\toprule
Experiment & Setting 
& Validity $\uparrow$ 
& Feature Value $\uparrow$ 
& Diversity $\uparrow$ \\
\midrule
\multirow{2}{*}{MorphoMNIST}
& Homogeneous
& $\phantom{+}1.0000_{\pm 0.0000}$
& $\phantom{+}0.3310_{\pm 0.0126}$
& $\phantom{+}0.0789_{\pm 0.0038}$ \\
& Diverse
& $+0.0000_{\pm 0.0000}$
& $+0.0119_{\pm 0.0003}$
& $+0.0206_{\pm 0.0010}$ \\
\midrule
\multirow{2}{*}{Crystal}
& Homogeneous
& $\phantom{+}1.0000_{\pm 0.0000}$
& $\phantom{+}8.4538_{\pm 0.0175}$
& $\phantom{+}14.9394_{\pm 0.2401}$ \\
& Diverse
& $+0.0000_{\pm 0.0000}$
& $-0.0730_{\pm 0.0198}$
& $+1.1998_{\pm 0.1015}$ \\
\bottomrule
\end{tabular}
}
\end{table}
\paragraph{Cross-pollination supports global exploration.}
As shown in \cref{tab:cross_pollination_aggregate}, diverse cross-pollination preserves perfect validity across both domains while consistently improving population diversity. 
On MorphoMNIST, diverse cross-pollination improves both thickness  and image-space diversity over the homogeneous baseline, with both metrics improving steadily across generations. 
The per-digit breakdown in \cref{tab:cross_pollination_per_digit} (Appendix \ref{app:more_results}) further supports this trend, with diverse cross-pollination maintaining perfect validity and improving diversity across all digits. 

On crystal data, diverse cross-pollination substantially increases latent diversity ($+1.2$) but yields a slight reduction in band gap ($-0.082$), reflecting a stronger exploration-exploitation tension in a more heterogeneous domain. 

\begin{figure}[h]
  \centering
  \includegraphics[width=0.95\linewidth]{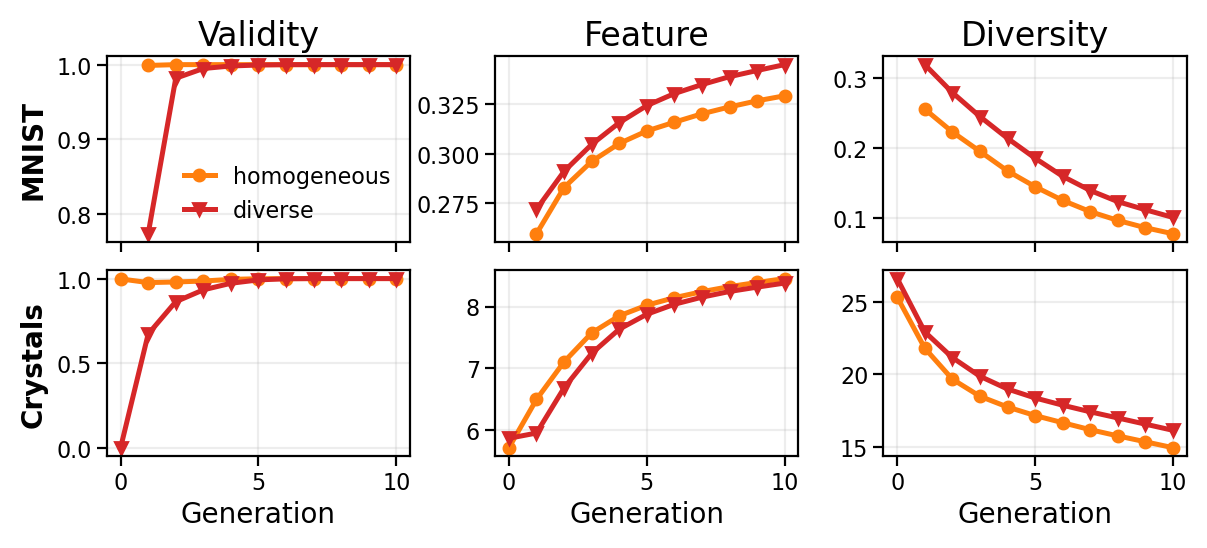}
  \caption{\textbf{Metrics across generations.} Rows: MorphoMNIST (top) and WyCryst (bottom) experiments. Colors show search regimes for cross-pollination. Columns show validity, feature value, and diversity. MorphoMNIST optimizes thickness and image diversity and WyCryst optimizes band gap and latent diversity.}
  \label{fig:cross-trend}
\end{figure}

Notably, \cref{fig:cross-trend} shows that diverse cross-pollination reaches target validity more slowly in the crystal setting, suggesting that residual combination across structurally different crystal systems introduces broader but slower targeted variation. Importantly, validity converges to the same level in both settings by generation 10, confirming that diverse cross-pollination expands the search space without sacrificing target conditioning.


\section{Discussion and Limitations}
\label{sec:discussion}

Our results demonstrate that the exploration-exploitation decomposition is an effective mechanism for residual-space generative editing across both image and scientific domains. Self-pollination and cross-pollination address different search regimes without sacrificing target validity. This is enabled by the residual structure exposed by conditional flow matching, which 
can be treated as the genotype for mutation, crossover, and selection, while the condition variable controls semantic attributes such as class identity. 

Beyond images, we demonstrate that the framework extends to crystal data, where conditions encode \textcolor{black}{the crystal system} and residual states provide a search space for evolutionary optimisation toward desired band gap properties. 
We apply cross-pollination to maximise the predicted band gap of crystal structures, demonstrating that the framework also supports scalar objectives. We note that the band gap is evaluated via a surrogate regressor rather than first-principles calculations, and that the physical stability of generated structures is not explicitly enforced; we therefore use this study as a demonstration of the optimisation mechanism rather than a novel materials discovery modeling approach. 
These findings indicate that any domain with an editable residual representation is a candidate for residual-space evolutionary optimisation, including molecular design and materials discovery.

Our current study is a controlled proof of concept, and several limitations remain. First, while we validate on both MorphoMNIST and crystal data, further validation on larger datasets and more complex generation tasks remains open. Second, key hyperparameters, including mutation strength, population size, selection criteria, and crossover design, require systematic ablation to assess robustness, efficiency, and failure modes. Addressing these questions is a promising direction for future work.

\bibliography{references}
\bibliographystyle{icml2026}

\newpage
\appendix
\onecolumn
\section{The Algorithm} \label{app:algorithm}
The complete algorithm is described in \Cref{alg:pollination}. 

\begin{algorithm}[ht]
\caption{Residual-Space Evolutionary Optimization}
\label{alg:pollination}
\begin{algorithmic}[1]
\REQUIRE Initial inputs $\mathcal{P}_0=\{x_j^{(0)}\}_{j=1}^{N}$, encoder $E$, decoder $D$, classifier $C$, frozen flow model $\flow$, target condition $\tgt$, generations $G$, population size $K$, child pool size $M$, mode $\in \{\mathrm{self},\mathrm{cross}\}$
\ENSURE Retained target-conditioned samples

\FOR{$g=1,\dots,G$}
    \STATE Initialize child set $\widetilde{\mathcal{P}}_g \leftarrow \emptyset$
    \STATE Encode current population: $z_j^{(g-1)} \leftarrow E(x_j^{(g-1)})$ for all $x_j^{(g-1)} \in \mathcal{P}_{g-1}$
    \STATE Predict source conditions: $\hat{c}_j^{(g-1)} \leftarrow \arg\max_c C(x_j^{(g-1)})_c$
    \STATE Lift current population:
    \[
        z_{\mathrm{res},j}^{(g-1)}
        \leftarrow
        \operatorname{Lift}\!\left(z_j^{(g-1)},\hat{c}_j^{(g-1)}\right)
    \]

    \FOR{$m=1,\dots,M$}
        \IF{$\mathrm{mode}=\mathrm{self}$}
            \STATE Sample one parent residual $z_{\mathrm{res},i}^{(g-1)}$
            \STATE Sample mutation noise $\epsilon^{(m)} \sim \mathcal{N}(0,\sigma^2 I)$
            \STATE Generate child residual:
            \[
                \tilde{z}_{\mathrm{res}}^{(m)}
                \leftarrow
                z_{\mathrm{res},i}^{(g-1)}+\epsilon^{(m)}
            \]
        \ELSE
            \STATE Sample two parent residuals $z_{\mathrm{res},i}^{(g-1)}$ and $z_{\mathrm{res},j}^{(g-1)}$
            \STATE Sample mixing weight $\alpha$ and mutation noise $\epsilon^{(m)}$
            \STATE Generate child residual:
            \[
                \tilde{z}_{\mathrm{res}}^{(m)}
                \leftarrow
                \operatorname{Crossover}\!\left(z_{\mathrm{res},i}^{(g-1)}, z_{\mathrm{res},j}^{(g-1)}, \alpha\right)
                +\epsilon^{(m)}
            \]
        \ENDIF

        \STATE Land child residual under target condition:
        \[
            \tilde{z}^{(m)}
            \leftarrow
            \operatorname{Land}(\tilde{z}_{\mathrm{res}}^{(m)},\tgt)
        \]
        \STATE Decode child sample: $\tilde{x}^{(m)} \leftarrow D(\tilde{z}^{(m)})$
        \STATE Add $\tilde{x}^{(m)}$ to $\widetilde{\mathcal{P}}_g$
    \ENDFOR

    \STATE Score all children in $\widetilde{\mathcal{P}}_g$ using $S_{\mathrm{self}}$ or $S_{\mathrm{cross}}$
    \STATE Select $K$ children as the next data population:
    \[
        \mathcal{P}_g
        \leftarrow
        \operatorname{Select}\!\left(\widetilde{\mathcal{P}}_g\right)
    \]
\ENDFOR
\STATE \textbf{return} final population $\mathcal{P}_G$
\end{algorithmic}
\end{algorithm}

\subsection{Lift--Land Operations}
\paragraph{Lift operation.}
Let $x$ denote an input image and $z=E(x)$ its autoencoder latent representation. The source condition is inferred from the classifier as
\[
\hat{c}=\arg\max_c C(x)_c .
\]
Given the conditional velocity field $\flow(z_t,t,c)$, the source-conditioned \emph{Lift} operation removes condition-specific information by integrating the flow backward from $t=1$ to $t=0$:
\[
\zres
=
\operatorname{Lift}(z,\hat{c}) .
\]
The resulting residual state $\zres$ is the representation used for mutation, crossover, and selection.

\paragraph{Land operation.}
Given a target condition $\tgt$, the \emph{Land} operation injects target-specific information by integrating the same flow forward from $t=0$ to $t=1$:
\[
z'=
\operatorname{Land}(\zres,\tgt).
\]
The edited image is obtained as
\[
x'=D(z').
\]

\paragraph{Leap step.}
A complete target-directed leap is the composition of Lift and Land:
\[
z'
=
\operatorname{Land}\!\left(
\operatorname{Lift}(E(x),\hat{c}),\tgt
\right),
\qquad
x'=D(z').
\]
Thus, the model first removes the source condition by lifting into residual space, and then imposes the target condition by landing under $\tgt$.

\subsection{Cross-pollination Crossover}
Given two parent residual states $z_{\mathrm{res},i}$ and $z_{\mathrm{res},j}$, cross-pollination constructs a child residual $\tilde{z}_{\mathrm{res}}$ by combining information from both parents, and then maps it back to the target-conditioned latent space through the Land operator.

\paragraph{Linear crossover.}
Linear crossover forms a convex interpolation between two parent residuals:
\[
\tilde{z}_{\mathrm{res}}
=
(1-\alpha) z_{\mathrm{res},i} + \alpha z_{\mathrm{res},j},
\]
where $\alpha \in [0,1]$ is the crossover mixing ratio. This operation produces a child residual that lies on the line segment between the two parents in residual space.

\paragraph{Dimension-wise crossover.}
Dimension-wise crossover combines two parent residuals coordinate by coordinate. Let $m \in \{0,1\}^{d}$ be a binary mask sampled independently for each residual dimension:
\[
m_k \sim \mathrm{Bernoulli}(\alpha),
\]
where $\alpha$ is the probability of inheriting dimension $k$ from the second parent. The child residual is
\[
\tilde{z}_{\mathrm{res},k}
=
(1-m_k) z_{\mathrm{res},i,k} + m_k z_{\mathrm{res},j,k},
\quad k = 1,\ldots,d.
\]
Equivalently,
\[
\tilde{z}_{\mathrm{res}}
=
(1-m) \odot z_{\mathrm{res},i}
+
m \odot z_{\mathrm{res},j},
\]
where $\odot$ denotes element-wise multiplication.

\subsection{Selection}
\paragraph{Top-$k$ selection.}
Given a candidate pool $\mathcal{C}=\{x'_i\}_{i=1}^{N}$ and a scalar selection score $S(x'_i)$ for each candidate, top-$k$ selection keeps the $K$ candidates with the largest scores:
\[
\mathcal{P}_{g}
=
\operatorname{TopK}_{x'_i \in \mathcal{C}}
\bigl(S(x'_i), K\bigr).
\]
This mode greedily preserves the highest-scoring candidates.

\paragraph{Tournament selection.}
Tournament selection repeatedly samples a small subset of candidates and keeps the best candidate within that subset. For each selected individual, we sample a tournament set
\[
\mathcal{T}_m \subset \mathcal{C},
\qquad |\mathcal{T}_m| = M.
\]
The winner is
\[
x'^{(m)}
=
\arg\max_{x'_i \in \mathcal{T}_m}
S(x'_i).
\]
After each winner is selected, it is removed from the remaining pool. Repeating this process $K$ times gives
\[
\mathcal{P}_{g}
=
\{x'^{(1)},\ldots,x'^{(K)}\}.
\]
Compared with top-$k$ selection, tournament selection introduces stochasticity while still favoring high-scoring candidates.


\section{Definitions}\label{app:definitions}
\paragraph{Target validity.}
For a generated image $x'$, the target-class probability used in the main text is
\[
p_{\mathrm{clf}}(\tgt \mid x')
=
\mathrm{softmax}\!\left(C(x')\right)_{\tgt}.
\]
The validity indicator is
\[
\mathbb{I}_{\mathrm{valid}}(x')
=
\mathbf{1}
\left[
\arg\max_c C(x')_c
=
\tgt
\right].
\]
For a generated population $\mathcal{P}=\{x'_i\}_{i=1}^{N}$, validity is reported as
\[
\mathrm{Validity}(\mathcal{P})
=
\frac{1}{|\mathcal{P}|}
\sum_{x'_i\in\mathcal{P}}
\mathbb{I}_{\mathrm{valid}}(x'_i).
\]

\paragraph{Classification margin.}
When used for selection, the classifier margin is
\[
m_{\tgt}(x')
=
C(x')_{\tgt}
-
\max_{c\neq \tgt} C(x')_c.
\]
A larger margin indicates that the classifier assigns the generated image more confidently to the target class.

\paragraph{Image similarity.}
For self-pollination, similarity measures how much the generated image preserves the source image. Given source image $x$ and edited image $x'$, we use
\[
\mathrm{sim}(x',x)
=
1
-
\frac{1}{HW}
\sum_{u=1}^{H}
\sum_{v=1}^{W}
\left(
 x'_{u,v} - x_{u,v}
\right)^2.
\]
For a population, we report the mean similarity over source--edit pairs. This definition matches the score $S_{\mathrm{self}}(x',x)$ in the main text.

\paragraph{Morphological feature value.}
For cross-pollination, we evaluate a MorphoMNIST feature
\[
\phi(x') \in
\{\mathrm{area}, \mathrm{length}, \mathrm{thickness}, \mathrm{slant}, \mathrm{width}, \mathrm{height}\}.
\]
By default, we use thickness, so the feature term in the main text is
\[
\mathrm{thickness}(x')=\phi(x').
\]
For a population $\mathcal{P}$, the mean feature value is
\[
\mathrm{MeanFeature}(\mathcal{P})
=
\frac{1}{|\mathcal{P}|}
\sum_{x'_i\in\mathcal{P}}
\phi(x'_i).
\]

\paragraph{Top-percentile feature value.}
For trend plots, we also report the mean feature value among the top feature percentile. Let $q_\rho$ be the $\rho$-th percentile of feature values in the population. The top-percentile feature metric is
\[
\mathrm{TopFeature}_{\rho}(\mathcal{P})
=
\frac{1}{|\mathcal{P}_{\rho}|}
\sum_{x'_i\in\mathcal{P}_{\rho}}
\phi(x'_i),
\]
where
\[
\mathcal{P}_{\rho}
=
\left\{
x'_i\in\mathcal{P}
:
\phi(x'_i) \ge q_\rho
\right\}.
\]
In our experiments, $\rho=95$ by default, so this reports the mean feature value of the top $5\%$ of the population.

\paragraph{Image diversity.}
In the main text, diversity denotes image-space diversity. We compute it as the mean pairwise angular distance between normalized flattened images:
\[
\mathrm{Diversity}(\mathcal{P})
=
\frac{1}{|\mathcal{P}|^2}
\sum_{x'_i,x'_j\in\mathcal{P}}
\left(
1-
\frac{\langle \operatorname{vec}(x'_i),\operatorname{vec}(x'_j)\rangle}
{\norm{\operatorname{vec}(x'_i)}_2\,\norm{\operatorname{vec}(x'_j)}_2}
\right).
\]

\paragraph{Crystal latent diversity.}
For the crystal experiments, diversity is measured in the crystal latent space rather than image space. Given a generated population $\mathcal{P}=\{z'_i\}_{i=1}^{|\mathcal{P}|}$, where each $z'_i$ is a crystal latent vector, we compute latent diversity as the mean pairwise Euclidean distance:
\[
\mathrm{Diversity}_{\mathrm{latent}}(\mathcal{P})
=
\frac{1}{|\mathcal{P}|^2}
\sum_{z'_i,z'_j\in\mathcal{P}}
\norm{z'_i-z'_j}_2 .
\]
This metric quantifies how broadly the generated candidates spread in the learned crystal latent space.

\paragraph{Self-pollination selection score.}
The main text uses the simplified score
\[
S_{\mathrm{self}}(x',x)
=
\mathrm{sim}(x',x)
+
\lambda\,p_{\mathrm{clf}}(\tgt\mid x').
\]
In the implementation, we additionally include a weight term:
\[
S_{\mathrm{self}}^{\mathrm{impl}}(x',x)
=
\lambda_{\mathrm{sim}}\mathrm{sim}(x',x)
+
\lambda_{\mathrm{conf}}p_{\mathrm{clf}}(\tgt\mid x')
+
\lambda_{\mathrm{margin}}m_{\tgt}(x').
\]

\paragraph{Cross-pollination selection score.}
The main text uses the simplified score
\[
S_{\mathrm{cross}}(x')
=
\mathrm{thickness}(x')
+
\lambda\,p_{\mathrm{clf}}(\tgt\mid x').
\]
In the implementation, we additionally include a weight term:
\[
S_{\mathrm{cross}}^{\mathrm{impl}}(x')
=
\lambda_{\mathrm{feat}}\phi(x')
+
\lambda_{\mathrm{conf}}p_{\mathrm{clf}}(\tgt\mid x')
+
\lambda_{\mathrm{margin}}m_{\tgt}(x').
\]

\section{Experiment Setup}
\label{app:experiment}
\subsection{Self-Pollination Experiment Setup}
\label{app:self_hyperparams}
The experiment evaluates all target digits $0,\ldots,9$. For each target digit,
source digits are randomly sampled from the non-target digits. We use $2048$
source samples per target digit, resulting in $20480$ evaluated source--target
edits in total. The underlying autoencoder, classifier, and conditional flow model are fixed during this experiment. The hyperparameters are shown in \Cref{tab:self_hyperparams}.

\begin{table}[ht]
\caption{Hyperparameters used for the self-pollination experiment.}
\label{tab:self_hyperparams}
\vskip 0.1in
\centering
\small
\begin{tabular}{ll}
\toprule
Hyperparameter & Value \\
\midrule
Random seed & $2023$,$2024$,$2025$ \\
Samples per target digit & $2048$ \\
Total evaluated edits & $20480$ \\
Leap step size & $0.3$ \\
Self-pollination generations & $10$ \\
Children per parent & $32$ \\
Mutation noise std. & $0.5$ \\
Selection mode & Top-$k$ \\
Target confidence weight $\lambda_\mathrm{conf}$ & $4.5$ \\
Similarity weight $\lambda_\mathrm{sim}$ & $1.5$ \\
Margin weight $\lambda_\mathrm{margin}$ & $0.5$ \\
\bottomrule
\end{tabular}
\end{table}

\subsection{Cross-Pollination Experiment Setup}
\label{app:cross_hyperparams}
\paragraph{MorphoMNIST}
The experiment evaluates all target digits $0,\ldots,9$. For each target digit,
we compare homogeneous and diverse cross-pollination. The homogeneous population
is initialized from samples of the target digit, whereas the diverse population
is initialized from randomly sampled non-target digits. 
population size. 
The underlying autoencoder, classifier, and conditional flow
model are fixed during this experiment. The hyperparameters are shown in
\Cref{tab:cross_hyperparams}.

\begin{table}[ht]
\caption{Hyperparameters used for the cross-pollination experiment.}
\label{tab:cross_hyperparams}
\vskip 0.1in
\centering
\small
\begin{tabular}{ll}
\toprule
Hyperparameter & Value \\
\midrule
Random seed & $2023$,$2024$,$2025$ \\
Optimized feature & Thickness \\
Crossover mode & Dimension \\
Population size per method & $512$ \\
Cross-pollination generations & $10$ \\
Children per parent & $32$ \\
Leap step size & $0.2$ \\
Crossover type & Dimension-wise crossover \\
Crossover mix ratio & $0.2$ \\
Mutation noise std. & $0.05$ \\
Selection mode & TopK \\
Target confidence weight $\lambda_\mathrm{conf}$ & $0.0$ \\
Feature weight $\lambda_\mathrm{feat}$ & $10.0$ \\
Margin weight $\lambda_\mathrm{margin}$ & $0.1$ \\
\bottomrule
\end{tabular}
\end{table}

\paragraph{Crystal structures.}
The crystal experiment evaluates all seven crystal systems:
Cubic, Hexagonal, Monoclinic, Orthorhombic, Tetragonal, Triclinic, and Trigonal.
For each target crystal system, we compare homogeneous and diverse
cross-pollination. The homogeneous population is initialized from structures
belonging to the target crystal system, whereas the diverse population is
initialized from structures sampled from non-target crystal systems. The
optimization objective is the predicted band gap. Before optimization, an
initial candidate pool is filtered by tournament selection using the true band
gap, and the selected candidates are used as the initial population. The
underlying crystal latent representation, crystal-system classifier, band-gap
regressor, and conditional flow model are fixed during this experiment. The
hyperparameters are shown in \Cref{tab:crystal_cross_hyperparams}.

\begin{table}[ht]
\caption{Hyperparameters used for the crystal cross-pollination experiment.}
\label{tab:crystal_cross_hyperparams}
\vskip 0.1in
\centering
\small
\begin{tabular}{ll}
\toprule
Hyperparameter & Value \\
\midrule
Random seed & $2023$,$2024$,$2025$ \\
Target crystal systems & $7$ systems \\
Optimized feature & Predicted band gap \\
Population size per method & $256$ \\
Cross-pollination generations & $10$ \\
Children per parent & $32$ \\
Leap step size & $0.2$ \\
Cross-pollination generations & $10$ \\
Crossover type & Dimension-wise crossover \\
Crossover mix ratio & $0.5$ \\
Mutation type & Gaussian residual mutation \\
Mutation noise std. & $0.05$ \\
Selection mode & TopK \\
Target confidence weight $\lambda_\mathrm{conf}$ & $0.5$ \\
Feature weight $\lambda_\mathrm{feat}$ & $10.0$ \\
Margin weight $\lambda_\mathrm{margin}$ & $0.1$ \\
Trend feature statistic & Top $5\%$ mean band gap \\
Diversity metric & Latent diversity \\
\bottomrule
\end{tabular}
\end{table}

\section{More Results} \label{app:more_results}
\begin{figure}[ht]
  \centering
  \includegraphics[width=0.95\linewidth]{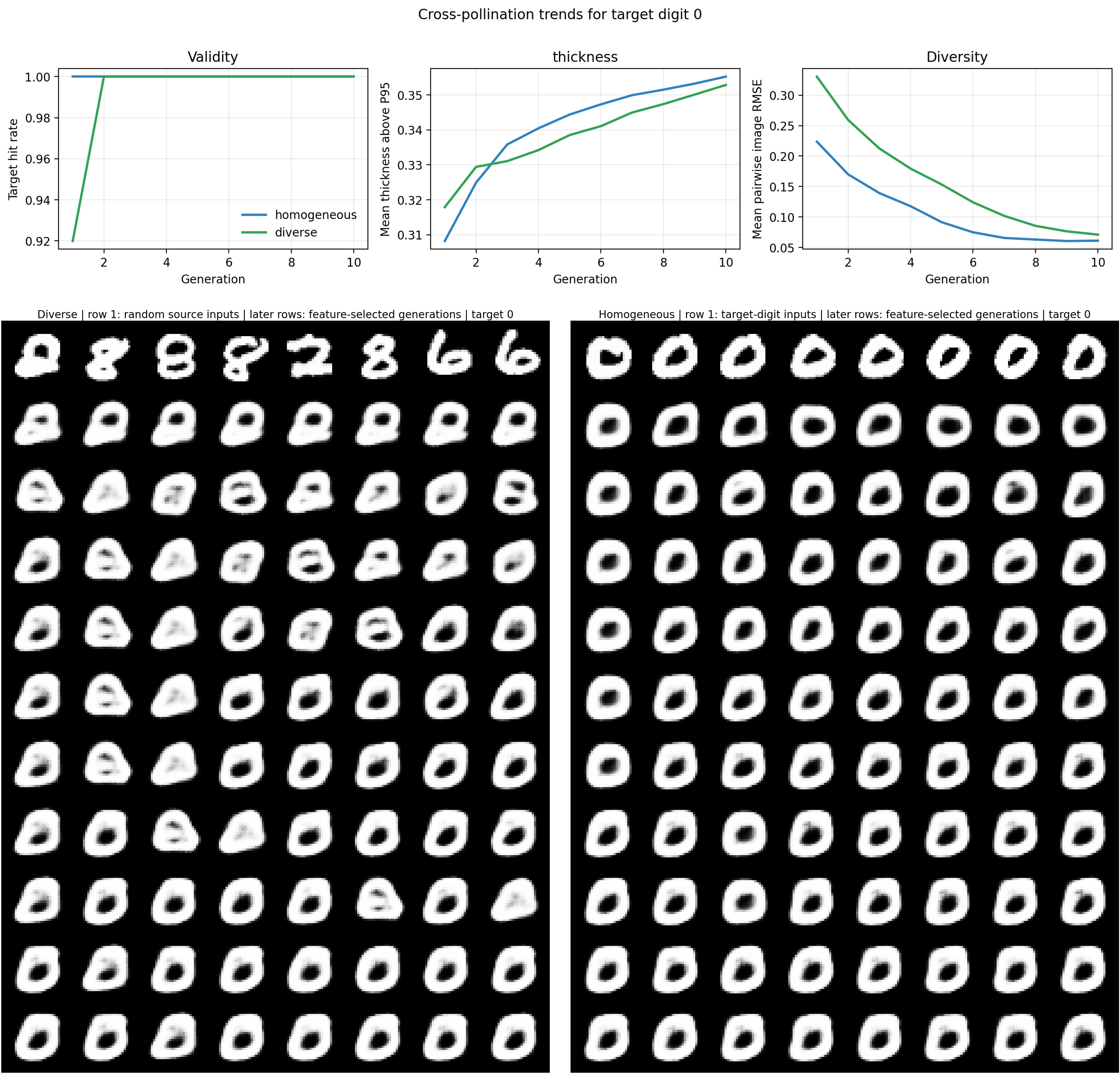}
  \caption{TBF}
  \label{fig:cross-demo-0}
\end{figure}

\begin{figure}[ht]
  \centering
  \includegraphics[width=0.95\linewidth]{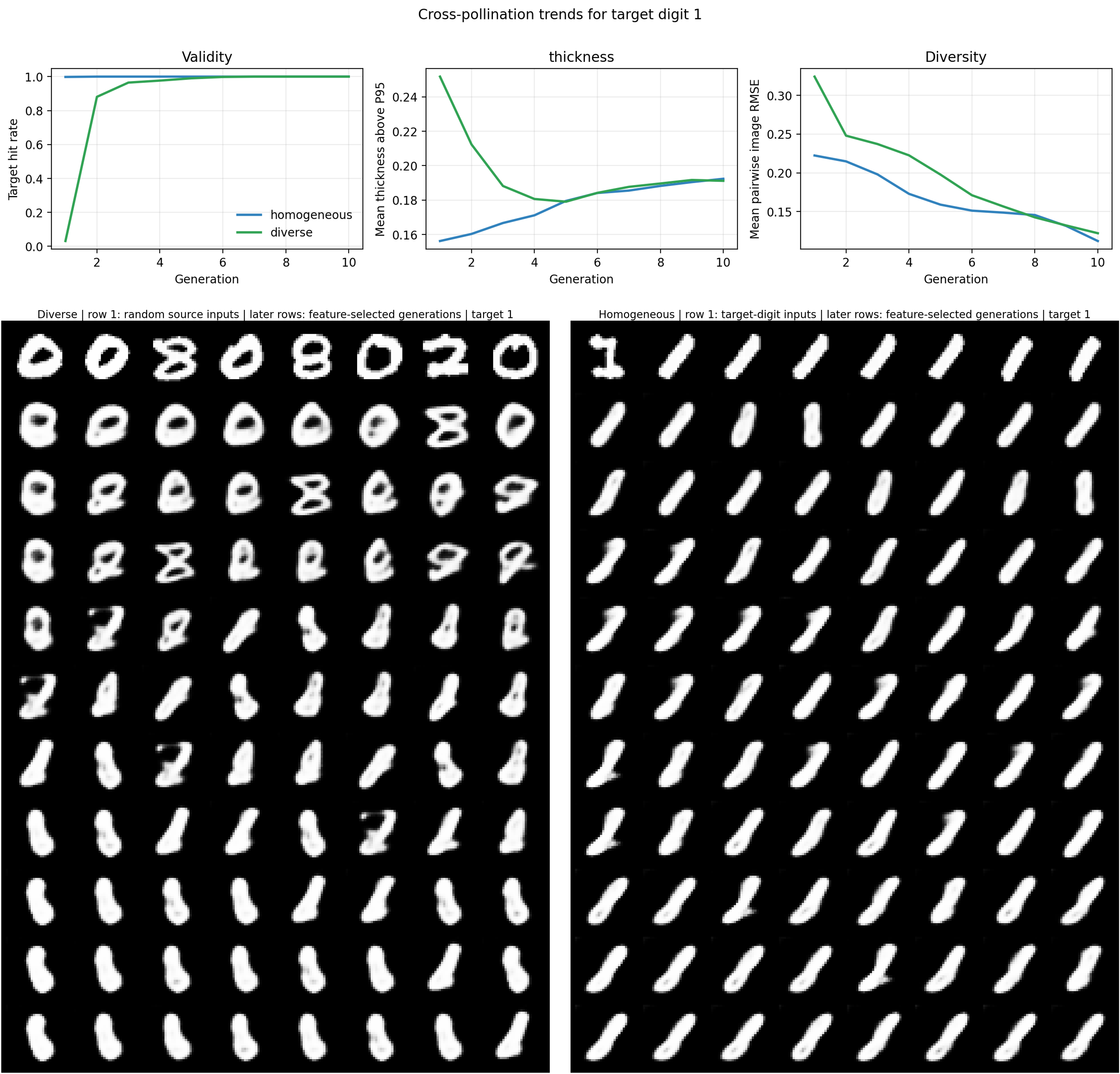}
  \caption{Similar as \Cref{fig:cross-demo-0} but for digit 1.}
  \label{fig:cross-demo-1}
\end{figure}

\begin{figure}[ht]
  \centering
  \includegraphics[width=0.95\linewidth]{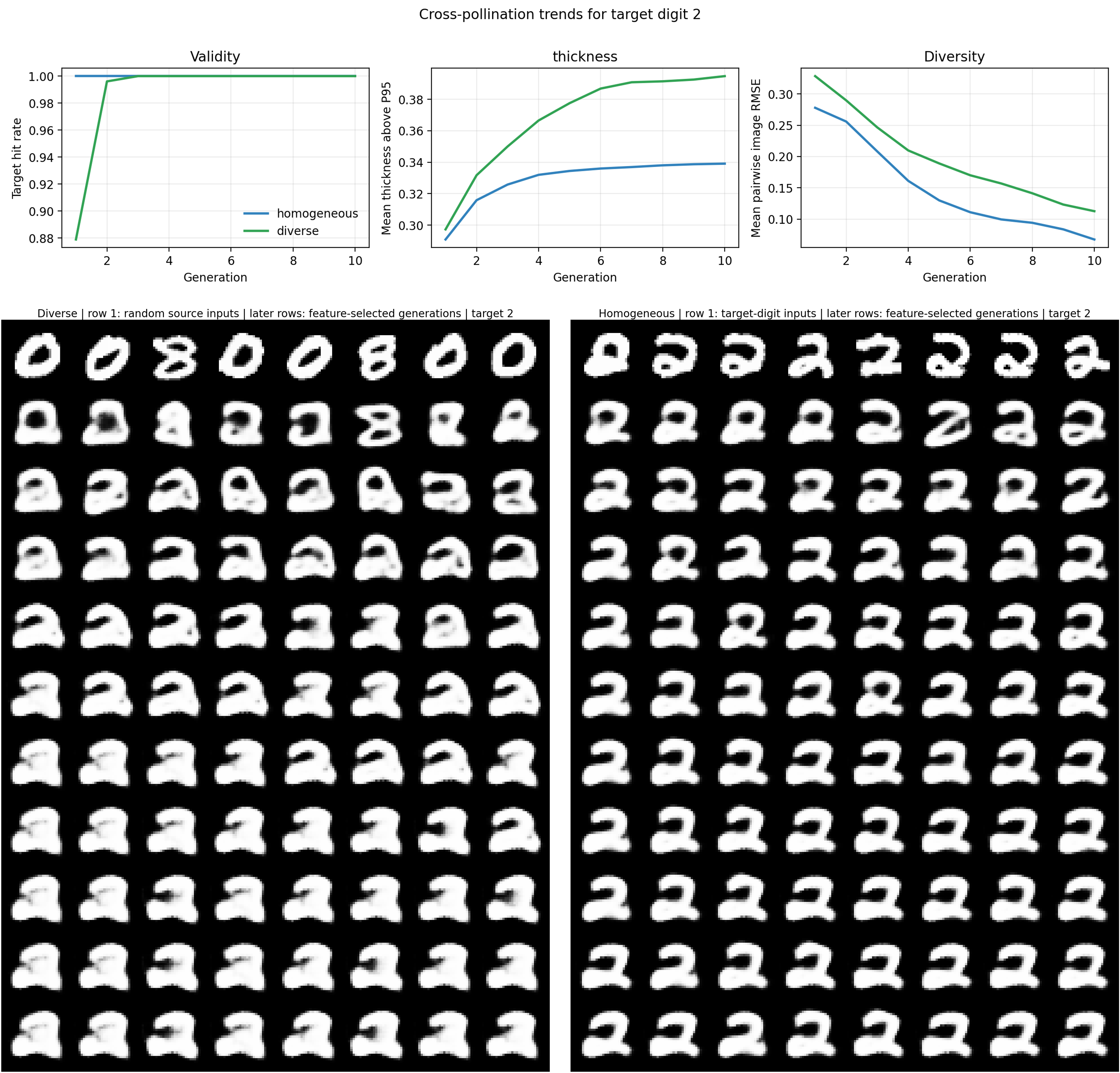}
  \caption{Similar as \Cref{fig:cross-demo-0} but for digit 2.}
  \label{fig:cross-demo-2}
\end{figure}

\begin{figure}[ht]
  \centering
  \includegraphics[width=0.95\linewidth]{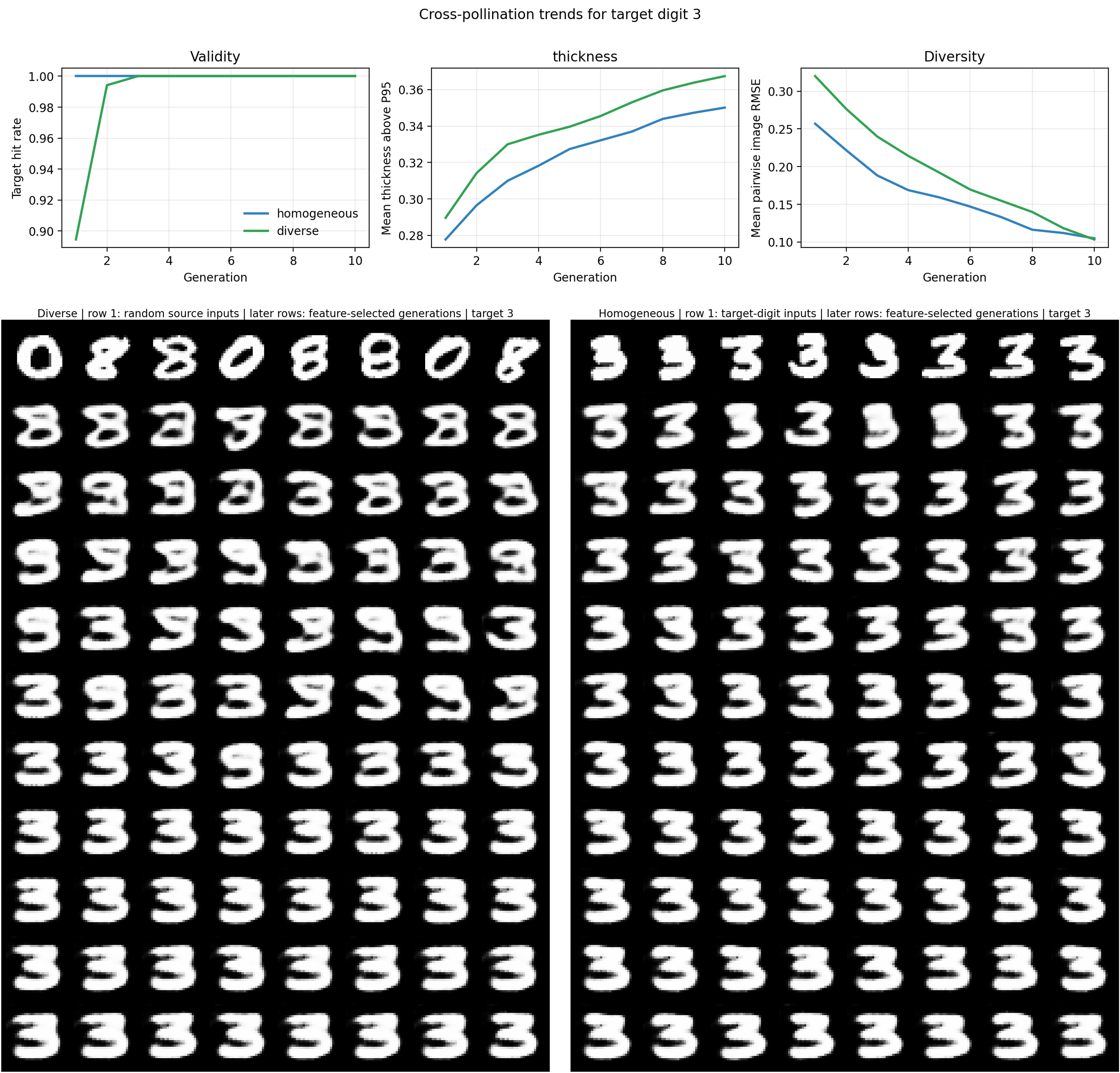}
  \caption{Similar as \Cref{fig:cross-demo-0} but for digit 3.}
  \label{fig:cross-demo-3}
\end{figure}

\begin{figure}[ht]
  \centering
  \includegraphics[width=0.95\linewidth]{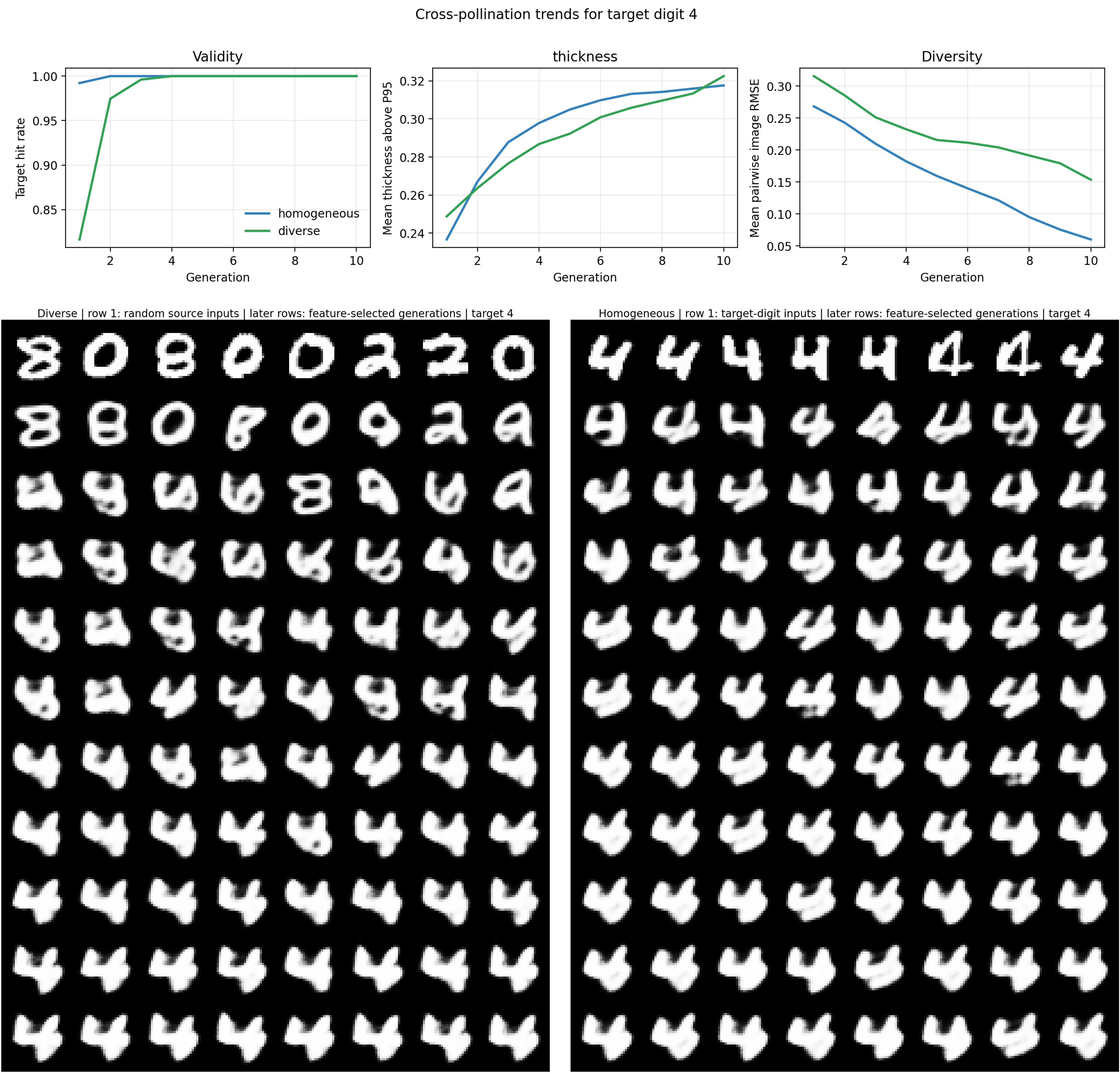}
  \caption{Similar as \Cref{fig:cross-demo-0} but for digit 4.}
  \label{fig:cross-demo-4}
\end{figure}

\begin{figure}[ht]
  \centering
  \includegraphics[width=0.95\linewidth]{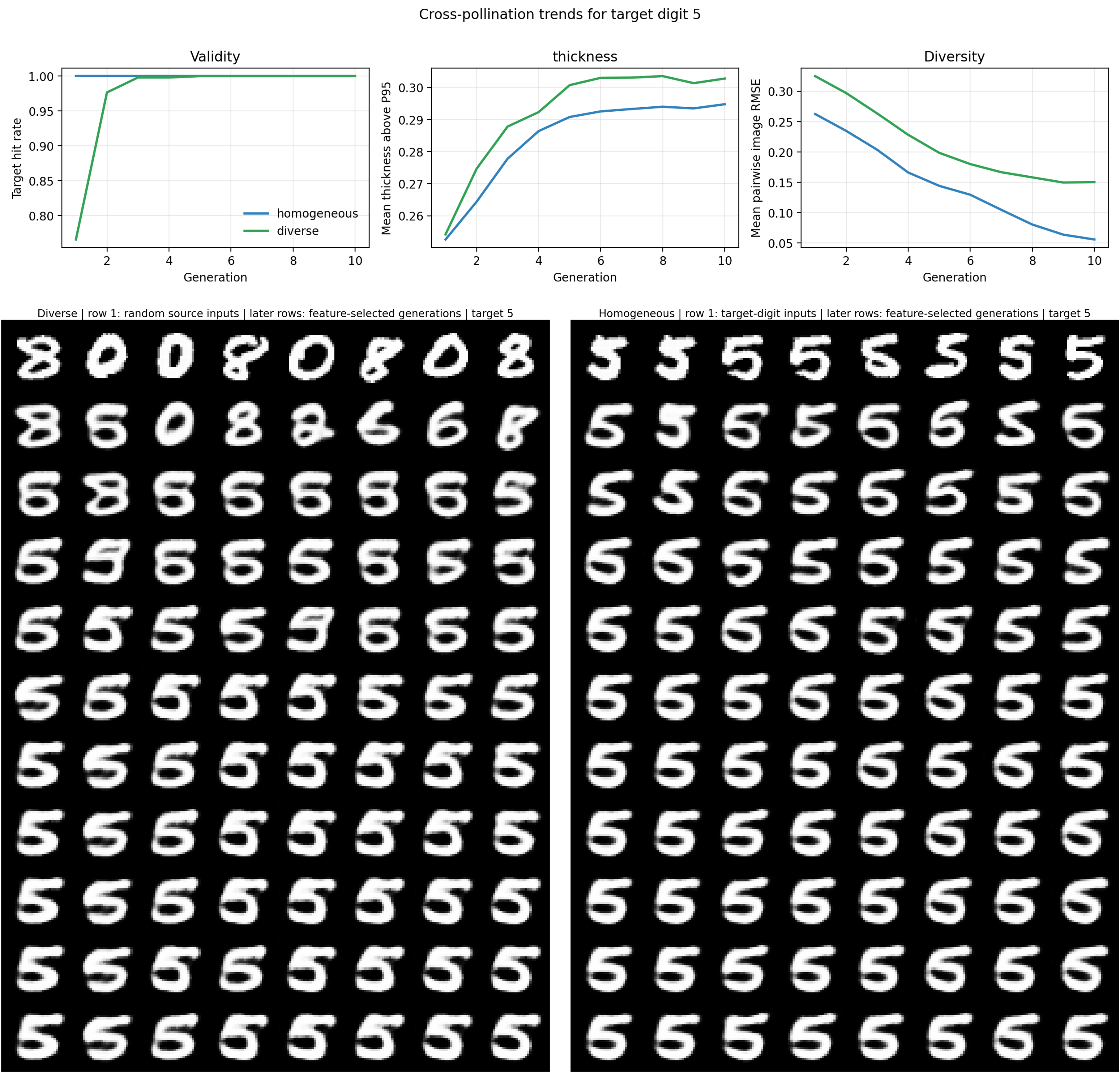}
  \caption{Similar as \Cref{fig:cross-demo-0} but for digit 5.}
  \label{fig:cross-demo-5}
\end{figure}

\begin{figure}[ht]
  \centering
  \includegraphics[width=0.95\linewidth]{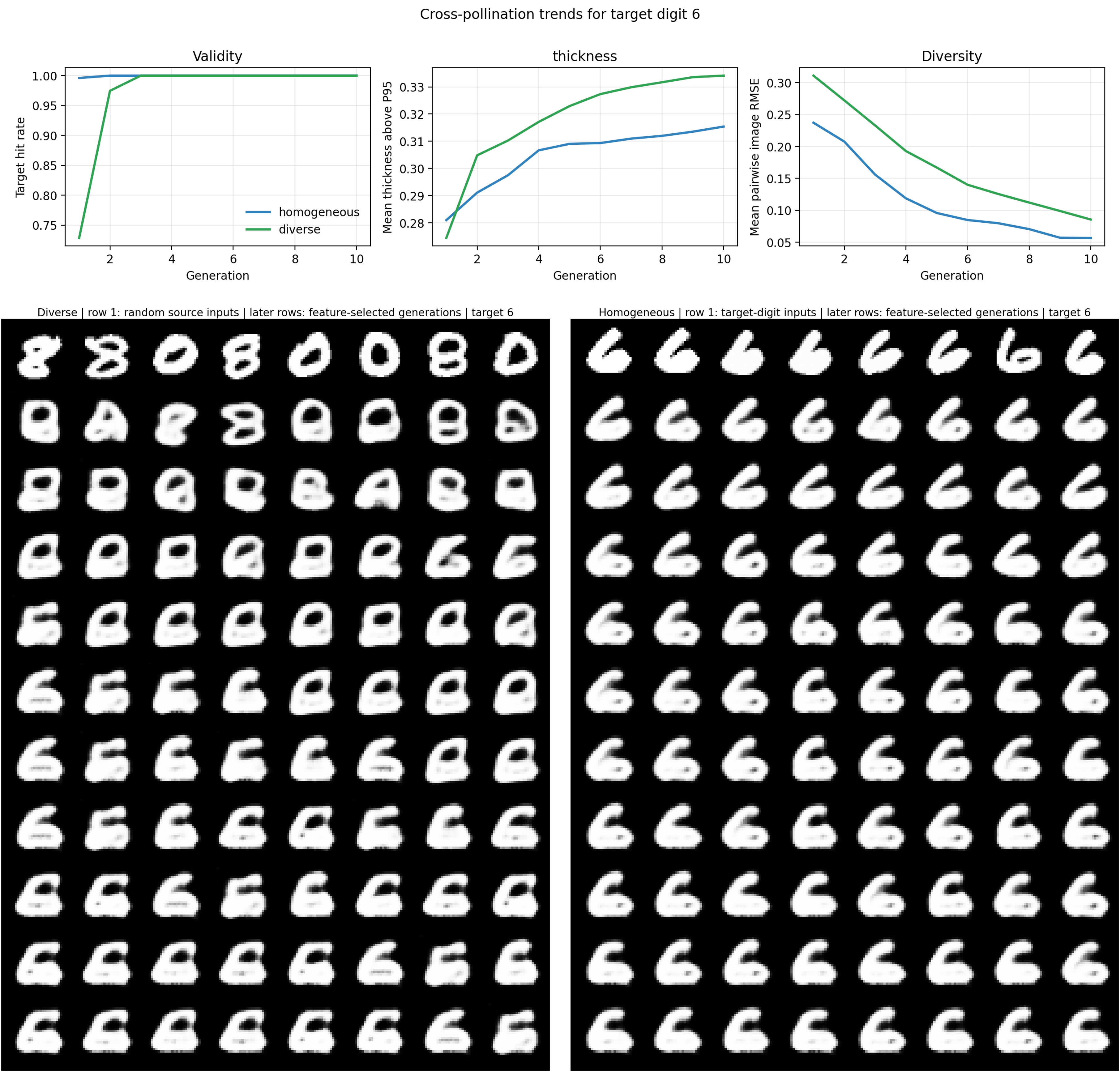}
  \caption{Similar as \Cref{fig:cross-demo-0} but for digit 6.}
  \label{fig:cross-demo-6}
\end{figure}

\begin{figure}[ht]
  \centering
  \includegraphics[width=0.95\linewidth]{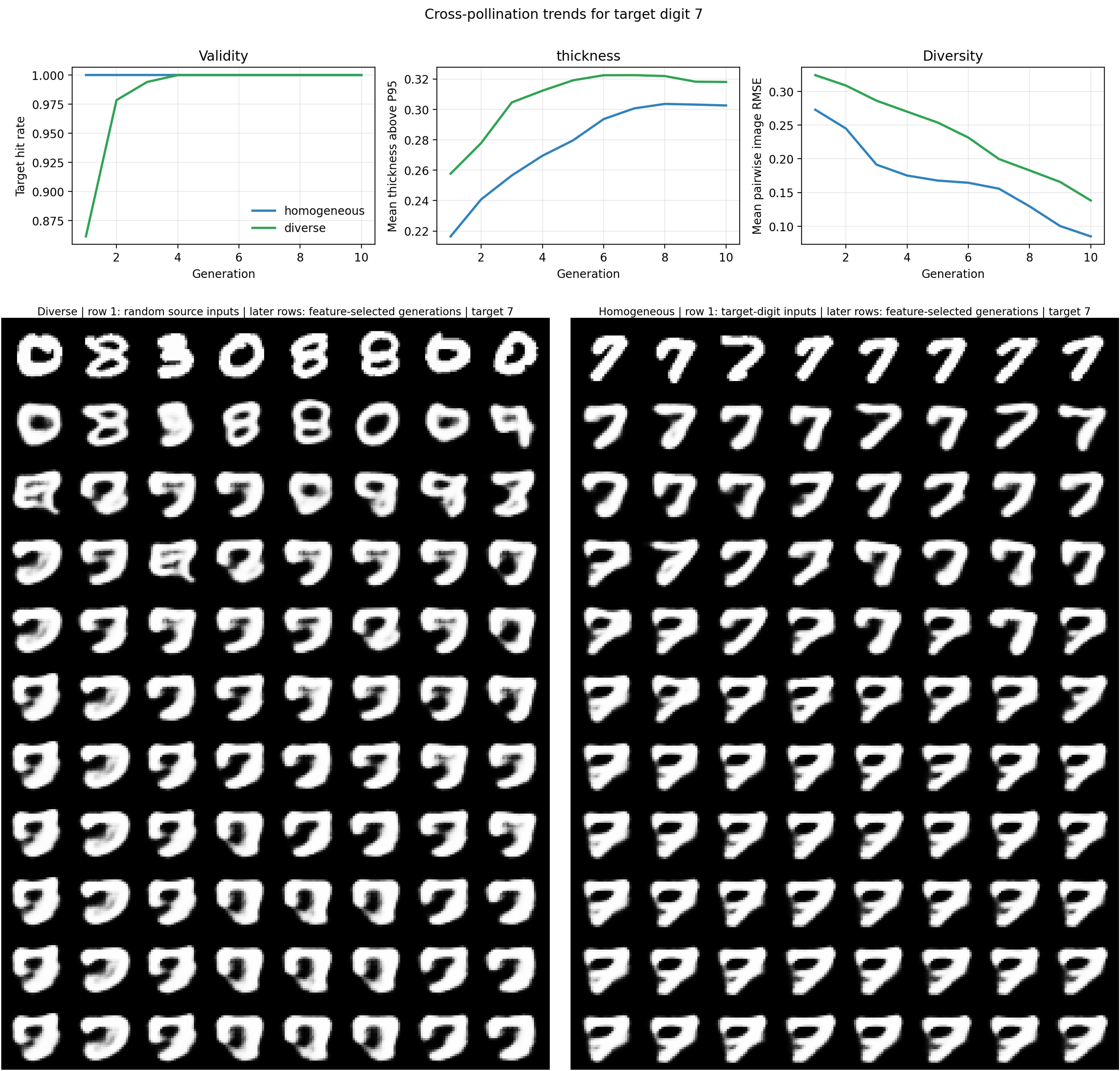}
  \caption{Similar as \Cref{fig:cross-demo-0} but for digit 7.}
  \label{fig:cross-demo-7}
\end{figure}

\begin{figure}[ht]
  \centering
  \includegraphics[width=0.95\linewidth]{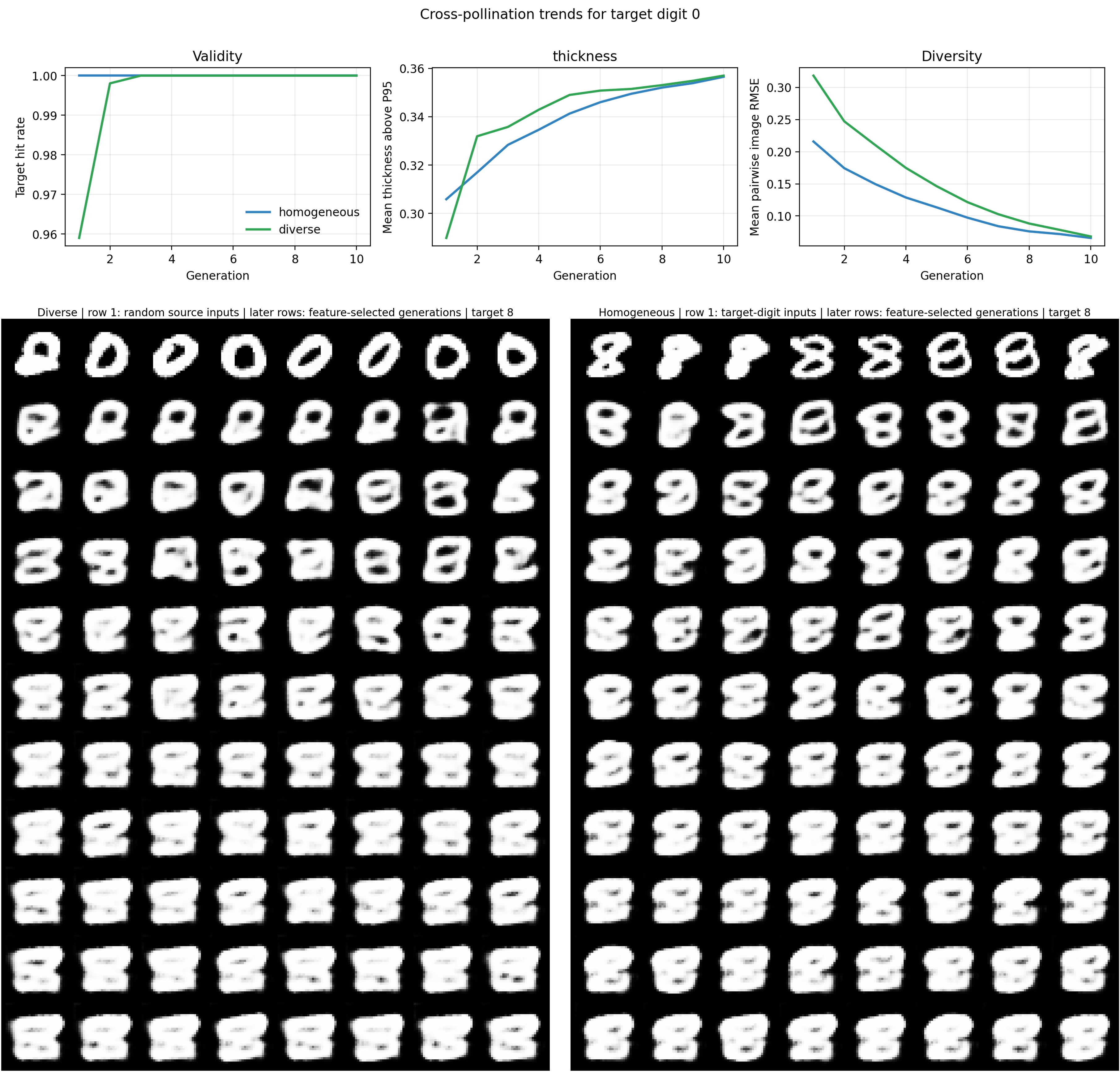}
  \caption{Similar as \Cref{fig:cross-demo-0} but for digit 8.}
  \label{fig:cross-demo-8}
\end{figure}

\begin{figure}[ht]
  \centering
  \includegraphics[width=0.95\linewidth]{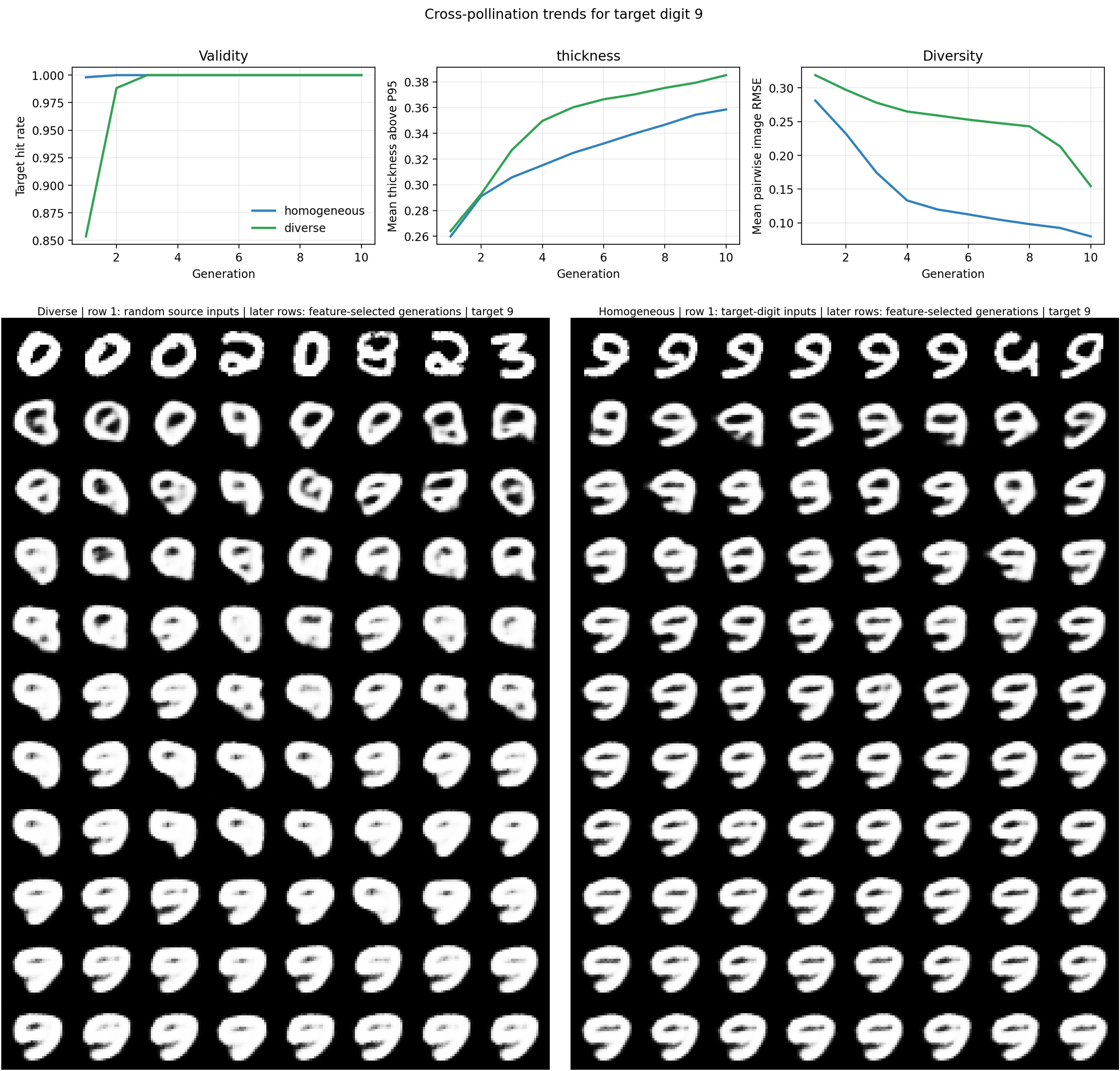}
  \caption{Similar as \Cref{fig:cross-demo-0} but for digit 9.}
  \label{fig:cross-demo-9}
\end{figure}

We present per-digit results for the self-pollination and cross-pollination experiments in \Cref{tab:per_digit_self_pollination,tab:cross_pollination_per_digit}.
Furthermore, we proved the per-digit result for each cross-pollination generation in \Cref{fig:cross-demo-0,fig:cross-demo-1,fig:cross-demo-2,fig:cross-demo-3,fig:cross-demo-4,fig:cross-demo-5,fig:cross-demo-6,fig:cross-demo-7,fig:cross-demo-8,fig:cross-demo-9}.

For WyCryst experiment, we provide per-target-crystal-system comparison in \Cref{tab:cross_pollination_per_crystal_system}.

\begin{table*}[t]
\centering
\caption{
Per-target-digit comparison between leap-only and self-pollination on MorphoMNIST.
Results are reported as mean$_{\pm \mathrm{STE}}$.
Self-pollination denotes the improvement over leap-only.
}
\label{tab:per_digit_self_pollination}
\scriptsize
\begin{tabular}{c|cc|cc}
\toprule
\multirow{2}{*}{Target Digit}
& \multicolumn{2}{c|}{Validity $\uparrow$}
& \multicolumn{2}{c}{Similarity $\uparrow$} \\
\cmidrule(lr){2-3}
\cmidrule(lr){4-5}
& Leap-only & Self-pollination
& Leap-only & Self-pollination \\
\midrule
0 
& $\phantom{+}0.9995_{\pm 0.0003}$ 
& $+0.0005_{\pm 0.0003}$ 
& $\phantom{+}0.9167_{\pm 0.0001}$ 
& $+0.0402_{\pm 0.0001}$ \\

1 
& $\phantom{+}0.9995_{\pm 0.0003}$ 
& $+0.0005_{\pm 0.0003}$ 
& $\phantom{+}0.9026_{\pm 0.0007}$ 
& $+0.0513_{\pm 0.0005}$ \\

2 
& $\phantom{+}0.9989_{\pm 0.0006}$ 
& $+0.0011_{\pm 0.0006}$ 
& $\phantom{+}0.9344_{\pm 0.0002}$ 
& $+0.0313_{\pm 0.0001}$ \\

3 
& $\phantom{+}0.9989_{\pm 0.0004}$ 
& $+0.0011_{\pm 0.0004}$ 
& $\phantom{+}0.9434_{\pm 0.0005}$ 
& $+0.0269_{\pm 0.0002}$ \\

4 
& $\phantom{+}0.9993_{\pm 0.0002}$ 
& $+0.0007_{\pm 0.0002}$ 
& $\phantom{+}0.9356_{\pm 0.0005}$ 
& $+0.0252_{\pm 0.0002}$ \\

5 
& $\phantom{+}0.9967_{\pm 0.0004}$ 
& $+0.0033_{\pm 0.0004}$ 
& $\phantom{+}0.9415_{\pm 0.0003}$ 
& $+0.0257_{\pm 0.0002}$ \\

6 
& $\phantom{+}1.0000_{\pm 0.0000}$ 
& $+0.0000_{\pm 0.0000}$ 
& $\phantom{+}0.9223_{\pm 0.0001}$ 
& $+0.0308_{\pm 0.0003}$ \\

7 
& $\phantom{+}0.9990_{\pm 0.0003}$ 
& $+0.0010_{\pm 0.0003}$ 
& $\phantom{+}0.9333_{\pm 0.0005}$ 
& $+0.0308_{\pm 0.0003}$ \\

8 
& $\phantom{+}0.9977_{\pm 0.0004}$ 
& $+0.0023_{\pm 0.0004}$ 
& $\phantom{+}0.9452_{\pm 0.0003}$ 
& $+0.0224_{\pm 0.0001}$ \\

9 
& $\phantom{+}0.9995_{\pm 0.0005}$ 
& $+0.0005_{\pm 0.0005}$ 
& $\phantom{+}0.9406_{\pm 0.0004}$ 
& $+0.0249_{\pm 0.0002}$ \\
\bottomrule
\end{tabular}
\end{table*}

\begin{table*}[t]
\centering
\caption{
Per-target-digit comparison between homogeneous and diverse cross-pollination.
Results are reported as mean$_{\pm \mathrm{STE}}$ over three random seeds.
Diverse denotes the improvement over homogeneous cross-pollination.
Validity denotes target-class success rate. Feature value corresponds to the optimized MorphoMNIST thickness value. Diversity denotes image-space diversity.
}
\label{tab:cross_pollination_per_digit}
\scriptsize
\setlength{\tabcolsep}{3.0pt}
\begin{tabular}{c|cc|cc|cc}
\toprule
\multirow{2}{*}{Target Digit}
& \multicolumn{2}{c|}{Validity $\uparrow$}
& \multicolumn{2}{c|}{Feature Value $\uparrow$}
& \multicolumn{2}{c}{Diversity $\uparrow$} \\
\cmidrule(lr){2-3}
\cmidrule(lr){4-5}
\cmidrule(lr){6-7}
& Homogeneous & Diverse
& Homogeneous & Diverse
& Homogeneous & Diverse \\
\midrule
0
& $\phantom{+}1.0000_{\pm 0.0000}$
& $+0.0000_{\pm 0.0000}$
& $\phantom{+}0.3568_{\pm 0.0009}$
& $-0.0028_{\pm 0.0009}$
& $\phantom{+}0.0608_{\pm 0.0032}$
& $+0.0117_{\pm 0.0034}$ \\

1
& $\phantom{+}1.0000_{\pm 0.0000}$
& $+0.0000_{\pm 0.0000}$
& $\phantom{+}0.1849_{\pm 0.0038}$
& $+0.0192_{\pm 0.0132}$
& $\phantom{+}0.0547_{\pm 0.0122}$
& $+0.0045_{\pm 0.0099}$ \\

2
& $\phantom{+}1.0000_{\pm 0.0000}$
& $+0.0000_{\pm 0.0000}$
& $\phantom{+}0.3432_{\pm 0.0010}$
& $+0.0327_{\pm 0.0033}$
& $\phantom{+}0.0702_{\pm 0.0010}$
& $+0.0305_{\pm 0.0103}$ \\

3
& $\phantom{+}1.0000_{\pm 0.0000}$
& $+0.0000_{\pm 0.0000}$
& $\phantom{+}0.3419_{\pm 0.0028}$
& $+0.0183_{\pm 0.0037}$
& $\phantom{+}0.0781_{\pm 0.0024}$
& $+0.0255_{\pm 0.0093}$ \\

4
& $\phantom{+}1.0000_{\pm 0.0000}$
& $+0.0000_{\pm 0.0000}$
& $\phantom{+}0.3132_{\pm 0.0014}$
& $+0.0086_{\pm 0.0060}$
& $\phantom{+}0.0714_{\pm 0.0045}$
& $+0.0315_{\pm 0.0089}$ \\

5
& $\phantom{+}1.0000_{\pm 0.0000}$
& $+0.0000_{\pm 0.0000}$
& $\phantom{+}0.2976_{\pm 0.0015}$
& $+0.0131_{\pm 0.0027}$
& $\phantom{+}0.0728_{\pm 0.0016}$
& $+0.0543_{\pm 0.0115}$ \\

6
& $\phantom{+}1.0000_{\pm 0.0000}$
& $+0.0000_{\pm 0.0000}$
& $\phantom{+}0.3208_{\pm 0.0002}$
& $+0.0151_{\pm 0.0025}$
& $\phantom{+}0.0835_{\pm 0.0063}$
& $+0.0170_{\pm 0.0067}$ \\

7
& $\phantom{+}1.0000_{\pm 0.0000}$
& $+0.0000_{\pm 0.0000}$
& $\phantom{+}0.2919_{\pm 0.0086}$
& $+0.0124_{\pm 0.0149}$
& $\phantom{+}0.1004_{\pm 0.0048}$
& $+0.0291_{\pm 0.0148}$ \\

8
& $\phantom{+}1.0000_{\pm 0.0000}$
& $+0.0000_{\pm 0.0000}$
& $\phantom{+}0.4803_{\pm 0.0015}$
& $+0.0104_{\pm 0.0110}$
& $\phantom{+}0.1026_{\pm 0.0043}$
& $+0.0160_{\pm 0.0050}$ \\

9
& $\phantom{+}1.0000_{\pm 0.0000}$
& $+0.0000_{\pm 0.0000}$
& $\phantom{+}0.3611_{\pm 0.0012}$
& $+0.0301_{\pm 0.0050}$
& $\phantom{+}0.0771_{\pm 0.0107}$
& $+0.0139_{\pm 0.0140}$ \\
\bottomrule
\end{tabular}
\end{table*}

\begin{table*}[t]
\centering
\caption{
Per-target-crystal-system comparison between homogeneous and diverse cross-pollination.
Results are reported as mean$_{\pm \mathrm{STE}}$ over three random seeds.
Diverse denotes the improvement over homogeneous cross-pollination.
Validity denotes target-crystal-system success rate. Band gap corresponds to the top-percentile mean predicted band gap. Diversity denotes latent-space diversity.
}
\label{tab:cross_pollination_per_crystal_system}
\scriptsize
\setlength{\tabcolsep}{3.0pt}
\begin{tabular}{c|cc|cc|cc}
\toprule
\multirow{2}{*}{Target System}
& \multicolumn{2}{c|}{Validity $\uparrow$}
& \multicolumn{2}{c|}{Band Gap $\uparrow$}
& \multicolumn{2}{c}{Latent Diversity $\uparrow$} \\
\cmidrule(lr){2-3}
\cmidrule(lr){4-5}
\cmidrule(lr){6-7}
& Homogeneous & Diverse
& Homogeneous & Diverse
& Homogeneous & Diverse \\
\midrule
Cubic
& $\phantom{+}1.0000_{\pm 0.0000}$
& $+0.0000_{\pm 0.0000}$
& $\phantom{+}8.2996_{\pm 0.0170}$
& $+0.0402_{\pm 0.0164}$
& $\phantom{+}15.8024_{\pm 0.4307}$
& $+0.3673_{\pm 0.5549}$ \\

Hexagonal
& $\phantom{+}1.0000_{\pm 0.0000}$
& $+0.0000_{\pm 0.0000}$
& $\phantom{+}8.4666_{\pm 0.0140}$
& $-0.0902_{\pm 0.0210}$
& $\phantom{+}14.9354_{\pm 0.5303}$
& $+1.3753_{\pm 0.2864}$ \\

Monoclinic
& $\phantom{+}1.0000_{\pm 0.0000}$
& $+0.0000_{\pm 0.0000}$
& $\phantom{+}8.4489_{\pm 0.0049}$
& $-0.0381_{\pm 0.0350}$
& $\phantom{+}15.2660_{\pm 0.5315}$
& $+1.6255_{\pm 0.9807}$ \\

Orthorhombic
& $\phantom{+}1.0000_{\pm 0.0000}$
& $+0.0000_{\pm 0.0000}$
& $\phantom{+}8.4305_{\pm 0.0345}$
& $-0.0705_{\pm 0.0093}$
& $\phantom{+}15.6388_{\pm 0.2830}$
& $+0.5593_{\pm 0.2099}$ \\

Tetragonal
& $\phantom{+}1.0000_{\pm 0.0000}$
& $+0.0000_{\pm 0.0000}$
& $\phantom{+}8.5034_{\pm 0.0142}$
& $-0.1105_{\pm 0.0203}$
& $\phantom{+}14.8050_{\pm 0.1352}$
& $+0.9827_{\pm 0.2945}$ \\

Triclinic
& $\phantom{+}1.0000_{\pm 0.0000}$
& $+0.0000_{\pm 0.0000}$
& $\phantom{+}8.4911_{\pm 0.0514}$
& $-0.1198_{\pm 0.0434}$
& $\phantom{+}12.7992_{\pm 0.5095}$
& $+2.6361_{\pm 0.3963}$ \\

Trigonal
& $\phantom{+}1.0000_{\pm 0.0000}$
& $+0.0000_{\pm 0.0000}$
& $\phantom{+}8.5366_{\pm 0.0187}$
& $-0.1220_{\pm 0.0570}$
& $\phantom{+}15.3291_{\pm 0.2934}$
& $+0.8526_{\pm 0.7583}$ \\
\bottomrule
\end{tabular}
\end{table*}

\end{document}